\documentclass[12pt]{article}
\usepackage{arxiv}
\usepackage[utf8]{inputenc}
\usepackage[T1]{fontenc}
\usepackage[greek,english]{babel}
\usepackage{amsmath}
\usepackage{hyperref}
\usepackage{url}
\usepackage{booktabs}
\usepackage{amsfonts}
\usepackage{graphicx}
\usepackage{subfigure}
\usepackage{natbib}
\usepackage{multirow,enumitem}
\usepackage{array}
\usepackage{algorithm}
\usepackage{authblk}
\usepackage[noend]{algpseudocode}

\newcolumntype{L}[1]{>{\raggedright\let\newline\\\arraybackslash\hspace{0pt}}m{#1}}
\newcolumntype{C}[1]{>{\centering\let\newline\\\arraybackslash\hspace{0pt}}m{#1}}
\newcolumntype{R}[1]{>{\raggedleft\let\newline\\\arraybackslash\hspace{0pt}}m{#1}}

\title{CHORUS: An Agentic Framework for Generating Realistic Deliberation Data}

\author[1]{Athanasios Koursaris}
\author[1]{George Domalis}
\author[1]{Alexandra Apostolopoulou}
\author[1]{Konstantinos Kanaris}
\author[1]{Dimitris Tsakalidis}
\author[1,2]{\href{https://orcid.org/0000-0002-3996-3301}{Ioannis E. Livieris}}

\affil[1]{Novelcore, Athens, GR 10436\\
	\texttt{\{koursaris,domalis,apostolopoulou,kanaris,tsakalidis\}@novelcore.eu}}

\affil[2]{Department of Business Administration \& Organization Administration\\
	 University of Peloponnese, Kalamata,
	 GR 24100 \texttt{livieris@uop.gr}}

\renewcommand{\shorttitle}{\normalsize{CHORUS: An Agentic Framework for Generating Realistic Deliberation Data}}

\hypersetup{
	pdftitle={CHORUS: An Agentic Framework for Generating Realistic Deliberation Data},
	pdfsubject={cs.CL, cs.AI},
	pdfauthor={Alexandra Apostolopoulou, Konstantinos Kanaris, Athanasios Koursaris, Dimitris Tsakalidis, George Domalis, Ioannis E. Livieris},
	pdfkeywords={Agentic AI, Large language models, Persona-based simulation, Deliberation data generation, Multi-agent simulation},
}

\begin{document}
    \maketitle

\begin{abstract}
Understanding the intricate dynamics of online discourse depends on large-scale deliberation data, a resource that remains scarce across interactive web platforms due to restrictive accessibility policies, ethical concerns and inconsistent data quality. In this paper, we propose \textsc{Chorus}, an agentic framework, which orchestrates LLM-powered actors with behaviorally consistent personas to generate realistic deliberation discussions. Each actor is governed by an autonomous agent equipped with memory of the evolving discussion, while participation timing is governed by a principled Poisson process-based temporal model, which approximates the heterogeneous engagement patterns of real users. The framework is further supported by structured tool usage, enabling actors to access external resources and facilitating integration with interactive web platforms. The framework was deployed on the \textsc{Deliberate} platform and evaluated by 30 expert participants across three dimensions: content realism, discussion coherence and analytical utility, confirming \textsc{Chorus} as a practical tool for generating high-quality deliberation data suitable for online discourse analysis. \\ \\
	*** This paper has been accepted for presentation at \textit{Engineering Applications and Advances of Artificial Intelligence 2026 (EAAAI'26) }. Cite: Koursaris, A., Domalis, G., Apostolopoulou, A., Kanaris, K., Tsakalidis, D., Livieris, I.E. (2026) CHORUS: An Agentic Framework for Generating Realistic Deliberation Data. In \emph{Engineering Applications and Advances of Artificial Intelligence}. ***
\end{abstract}

% keywords can be removed
\keywords{Agentic AI \and Large language models \and Persona-based simulation \and Deliberation data generation \and Multi-agent simulation.}

\section{Introduction}
Deliberation data encompasses all forms of structured and unstructured records in the context of online participatory discourse, including argumentative exchanges, opinion expressions and interactive responses, generated within web platforms designed to facilitate rational and inclusive discussion among citizens \citep{behrendt2025natural}. Such data constitute a critical resource for online discourse analysis, underpinning objectives ranging from studying deliberation dynamics and communication patterns to benchmarking natural language processing pipelines \citep{lawrence2019argument,steenbergen2003measuring}. However, the acquisition of deliberation data at sufficient scale poses a significant challenge, exacerbated by platform access restrictions, ethical considerations surrounding user privacy and inconsistent quality of available datasets \citep{argyle2023out,bisbee2023synthetic}. As the demand for large-scale public discourse analysis grows across research, policy-making and digital platform governance, the problem of deliberation data scarcity becomes increasingly acute, underscoring the need for alternative approaches capable of generating high-quality, realistic and ethically sound deliberation data.

Although simulation-driven approaches have demonstrated remarkable capabilities in modeling complex interactive settings \citep{park2023generative,yang2024oasis}, their application to deliberation data generation remains constrained by several fundamental limitations. Most notably, such frameworks exhibit a predominant focus on goal-directed tasks, rather than the open-ended, multi-turn deliberative discourse characteristic of authentic participatory exchanges \citep{liu2023agentbench,wang2025agenta}. Existing approaches incorporating user personas have been primarily developed for security testing and software evaluation \citep{lu2025uxagent,mayr2025chatchecker}, remaining limited in producing sustained, contextually coherent conversations and emulating the emergent multi-user dynamics observed during authentic interaction \citep{casoria2025evaluating,li2024steerability}. Furthermore, the heterogeneous temporal patterns shaping internet discourse are largely unaddressed, as current solutions lack principled mechanisms for modeling the diverse participation rates characteristic of real platforms. These shortcomings underscore the need for frameworks combining behavioral fidelity, temporal realism and contextual awareness to adequately support deliberative environments.

To address these challenges, we propose \textsc{Chorus}\footnote{The name \textsc{Chorus} is inspired by the Chorus of ancient Greek tragedy, a collective of voices that commented on, reacted to and shaped the unfolding narrative, mirroring the role of simulated actors in participatory discourse} (CHaracter-driven Orchestrated Response \& User Simulation), an agentic simulation framework able to generate realistic deliberation data by orchestrating LLM-powered actors on interactive web platforms. \textsc{Chorus} grounds each actor's behavior in a structured persona and equips it with contextual awareness of the evolving discussion, producing dialogue which is both semantically coherent and behaviorally diverse. A principled Poisson process-based model governs the temporal dynamics of actor participation, while structured tools facilitate web platform integration and interactivity, providing actors with the ability to access external resources. Each actor is governed by an autonomous agent equipped with memory of the evolving discussion and a persona-driven behavioral profile. \textsc{Chorus} was deployed on \textsc{Deliberate} and evaluated across three dimensions: content realism, discussion coherence and analytical utility, confirming that it constitutes a practical tool for generating high-quality deliberation data suitable for online discourse analysis.

\begin{itemize}
	\item We develop an agentic simulation framework named \textsc{Chorus}, which orchestrates a cast of LLM-powered actors with clearly defined, behaviorally consistent personas to generate realistic deliberation discussions on interactive web platforms.
	
	\item We propose a Poisson process-based temporal model governing actor participation dynamics, alongside a structured tool suite enabling seamless integration with external web platforms through the translation of generative outputs into concrete platform actions.
	
	\item We deploy \textsc{Chorus} on the public platform, \textsc{Deliberate}, and evaluate its performance across content realism, discussion coherence and analytical utility, demonstrating its effectiveness as a practical tool for generating high-quality deliberation data without reliance on real user participation.
\end{itemize}

The remainder of this paper is organized as follows. Section~\ref{Sec:2} reviews related work on agentic simulation frameworks and persona-based approaches. Section~\ref{sec:3} presents the \textsc{Chorus} framework, detailing its actor model, temporal dynamics and tool integration. Section~\ref{Sec:5} describes the deployment on \textsc{Deliberate}, reports the generated discussion outputs and presents the expert evaluation results. Finally, Section~\ref{Sec:6} 
concludes the paper and outlines directions for future research.

\section{Related work}\label{Sec:2}

The development of effective simulation frameworks for interactive web platforms represents a confluence of two fundamental research directions: (1) agentic orchestration architectures, which coordinate multiple LLM-powered agents to simulate authentic user-to-user interactions through shared platforms and (2) persona-based simulation strategies, which generate behaviorally realistic synthetic users. The first direction focuses on multi-agent simulation, including orchestration capabilities, coordination mechanisms and the social dynamics necessary for realistic multi-user interaction. The second direction concerns the construction of behaviorally diverse agent personas, including critical work on data-driven behavioral grounding. In the rest of this section, we provide a comprehensive review across these areas, highlighting their contributions and limitations which motivate the development of \textsc{Chorus}.

\cite{park2023generative} deployed 25 LLM-powered agents in ``Smallville'', each equipped with observation, memory with reflection and planning modules, demonstrating emergent capabilities such as autonomously spreading event invitations, forming relationships and coordinating group activities. The architecture established that LLM-powered agents grounded in memory and planning mechanisms can produce believable social behaviors without explicit scripting. However, the framework remains isolated from real interactive platforms,
where generated content must maintain semantic coherence with an ongoing
and dynamically evolving deliberation.

\cite{sun2025persona} introduced Persona-L, leveraging LLMs and an ability-based framework to generate personas of people with complex needs, demonstrating that careful prompting strategies can expand persona diversity beyond typical demographic representations. The methodology produces synthetic user profiles capturing nuanced combinations of abilities, preferences and behavioral characteristics, though it addresses persona construction in isolation without modeling how multiple concurrent personas collectively shape discourse within shared interactive environments.

\cite{yang2024oasis} scaled agentic simulation with OASIS, supporting up to one million LLM-based agents on simulated social media platforms, replicating phenomena including information spreading, group polarization and herd effects. The authors demonstrated that larger agent group scale leads to enhanced group dynamics and more realistic emergent collective behaviors, establishing scale as a critical factor in simulation fidelity. Despite these advances, the framework remains oriented toward modeling broad social dynamics rather than generating the semantically coherent, multi-turn deliberative discourse required for structured participatory platforms.

\cite{li2024steerability} demonstrated that demographic-based personas exhibit limited predictive accuracy compared to data-driven behavioral patterns, with substantial performance improvements when steering LLMs toward behavioral patterns extracted from actual user response data. This finding motivates the integration of empirical behavioral grounding into persona design, though the work does not extend to modeling temporal participation dynamics or multi-party discourse structures characteristic of real platform activity.

\cite{casoria2025evaluating} found that model censorship impacts the ability of LLM-generated personas to capture controversial or ``negative'' traits such as low agreeableness or high neuroticism, reducing behavioral diversity in synthetic user populations. This limitation is particularly relevant for deliberation simulation, where skeptical, contrarian, or disengaged user archetypes are essential for generating discussions that approximate the full spectrum of authentic participatory discourse.

Despite the advances reported across these two research directions, their applicability to realistic deliberation data generation remains limited. Existing agentic frameworks simulate social dynamics within self-contained virtual environments without interfacing with real platforms, while persona-based methods enhance behavioral realism but do not model how concurrent participants collectively shape discourse over time. A further limitation shared across both lines of work is the absence of principled temporal modeling, as the participation patterns and engagement asymmetries characterizing organic platform activity remain outside the scope of current approaches. \textsc{Chorus} bridges these gaps by deploying persona-driven agents directly on interactive platforms within a unified orchestration framework, coupling behavioral consistency with a stochastic formulation of participation dynamics to produce realistic deliberation data exhibiting both the content diversity and temporal structure of authentic user interaction.

\section{\textsc{Chorus}: Proposed Framework}\label{sec:3}

The primary objective of \textsc{Chorus} is the generation of realistic deliberation data through the interaction of independent actors. It instantiates a set of $N$ actors $\mathcal{A} = \{a_1, \ldots, a_N\}$, each governed by an autonomous agent, whose interactions are orchestrated. Each actor $a_i \in \mathcal{A}$, with $i=1,2,\dots N$ is characterized by a persona $\rho_i$, which encapsulates its biographical context, communication style, core beliefs and engagement patterns, serving as the behavioral blueprint that constrains the underlying agent's outputs to remain consistent with a specific user archetype throughout the simulation. The agent associated with each actor maintains awareness of the evolving discussion by continuously accessing its own personal posting and actions history, $\mathcal{H}_i^{\textnormal{post}}$ and $\mathcal{H}_i^{\textnormal{action}}$, respectively, producing actions and posts that reflect both its persona-driven identity and the current state of the deliberation as well as the shared history $\mathcal{H}$. Specifically, $\mathcal{H}_i^{\textnormal{post}}$ records all content previously 
submitted by actor $a_i$, while $\mathcal{H}_i^{\textnormal{action}}$ tracks all 
voting actions performed, preventing redundant engagement with already-acted-upon content.

Algorithm~\ref{alg:main} presents the main simulation cycle governing the execution of the agentic deliberation framework. A key design decision concerns the mechanism governing when each actor participates in the discussion. A persona-driven approach, wherein the agent itself decides whether to post or act based on its persona and the current discussion state, was deliberately avoided, as it introduces systematic bias: the LLM's intrinsic tendencies would disproportionately shape participation timing, resulting in engagement patterns that reflect model priors rather than realistic user behavior \citep{casoria2025evaluating,li2024steerability}. Instead, the temporal dynamics of each actor's participation are governed by 
independent Poisson processes parameterized by actor-specific rates 
$\lambda_i^{\textnormal{post}}$ and $\lambda_i^{\textnormal{action}}$, introducing principled 
stochasticity that approximates the heterogeneous engagement patterns observed 
on real platforms. Intuitively, $\lambda_i^{\textnormal{post}}$ and $\lambda_i^{\textnormal{action}}$ represent the expected number of posts and actions performed by actor $a_i$ per unit time, respectively; higher values encode more active engagement profiles.

The algorithm initializes a global priority queue $\mathcal{Q}$ and for each actor $a_i$, samples initial posting and action event times from $\mathrm{Poisson}(\lambda_i^{\textnormal{post}})$ and $\mathrm{Poisson}(\lambda_i^{\textnormal{action}})$ respectively, inserting the corresponding events into $\mathcal{Q}$. The simulation proceeds by iteratively extracting the earliest pending event from $\mathcal{Q}$ and dispatching the associated procedure until either the queue is exhausted or the simulation horizon $T$ is reached. Upon termination, the algorithm returns $\mathcal{H} = \bigcup_{i=1}^N \left(\mathcal{H}_i^{\textnormal{post}} \cup \mathcal{H}_i^{\textnormal{action}}\right)$, the complete shared discussion history. Thus, $\mathcal{H}$ constitutes the complete record of all synthetic deliberation 
activity, serving as the primary output of the simulation and the input to 
downstream NLP analytical pipelines.

Algorithm~\ref{alg:action} presents the \textsc{Action} procedure, which governs the voting behavior of actor $a_i$. The agent reasons over the aggregated post history $\bigcup_{j=1}^N \mathcal{H}_j^{\textnormal{post}}$ and its personal action history $\mathcal{H}_i^{\textnormal{action}}$ to select a set of $M$ candidate posts $\mathcal{C}_i$ for potential voting, leveraging the actor's persona $\rho_i$ to ensure behavioral consistency. For each candidate post $c \in \mathcal{C}_i$, a Bernoulli trial is performed by drawing $u \sim \mathcal{U}(0,1)$; the vote is executed only if $u$ exceeds the actor-specific threshold $\theta_i^{\textnormal{action}}$, introducing controlled stochasticity in action execution. Each executed vote is recorded in $\mathcal{H}_i^{\textnormal{action}}$, ensuring that actors do not repeatedly act on the same content. Intuitively, $\theta_i^{\textnormal{action}}$ acts as a selectivity filter: actors with higher thresholds engage more discriminately with existing content, while lower values encode indiscriminate engagement behavior.

\begin{algorithm}[!ht]
	\caption{Agentic Simulation Cycle}\label{alg:main}
	\begin{algorithmic}[1]
		\Statex \textbf{Parameters:}
		\Statex \hspace{0.3cm}$T$: simulation duration
		\Statex \hspace{0.3cm}$\mathcal{A} = \{a_1, \ldots, a_N\}$: actor set
		\Statex \hspace{0.3cm}$\lambda_i^{\textnormal{post}}, \lambda_i^{\textnormal{action}} > 0$: Poisson rates of the $i$-th actor $a_i$, $\forall i=1,2,\dots,N$
		\Statex \textbf{Output:}
		\Statex \hspace{0.3cm}$\mathcal{H} = \bigcup_{i=1}^N \left(\mathcal{H}_i^{\textnormal{post}} \cup \mathcal{H}_i^{\textnormal{action}}\right)$: shared discussion history
		\vspace{.2cm}    
		\State $\mathcal{Q} \leftarrow \emptyset$\Comment{Initialization}
		\For{$i = 1$ \textbf{to} $N$}
		\State $\mathcal{H}_i^\textnormal{post} \leftarrow \emptyset$;\; $\mathcal{H}_i^\textnormal{action} \leftarrow \emptyset$
		\State $t_i^{\textnormal{post}} \sim \mathrm{Poisson}(\lambda_i^{\textnormal{post}})$;\; $\mathcal{Q}.\textsc{Push}(t_i^{\textnormal{post}},\, a_i,\, \textsc{Post})$
		\State $t_i^{\textnormal{action}} \sim \mathrm{Poisson}(\lambda_i^{\textnormal{action}})$;\; $\mathcal{Q}.\textsc{Push}(t_i^{\textnormal{action}},\, a_i,\, \textsc{Action})$
		\EndFor
		\While{$\mathcal{Q} \neq \emptyset$}
		\State $(t^*,\, a_i,\, \textsc{Proc}) \leftarrow \mathcal{Q}.\textsc{Pop}()$ \Comment{Extract earliest event}
		\State \textbf{If} {$t^* > T$} \textbf{break}
		\State $\textsc{Proc}(a_i)$ \Comment{Execute $\textsc{Action}(a_i)$ or $\textsc{Post}(a_i)$}
		\State $t^{\textnormal{next}} \leftarrow t^* + \mathrm{Poisson}(\lambda_i^{\textsc{Proc}})$;\; $\mathcal{Q}.\textsc{Push}(t^{\textnormal{next}},\, a_i,\, \textsc{Proc})$
		\EndWhile
		\State \Return $\mathcal{H}$    
	\end{algorithmic}
\end{algorithm}

\begin{algorithm}[!ht]
	\caption{$\textsc{Action}(a_i)$}\label{alg:action}
	\begin{algorithmic}[1]
		\Statex \textbf{Parameters:}
		\Statex \hspace{0.3cm}$\rho_i$: Persona of the $i$-th actor $a_i$
		\Statex \hspace{0.3cm}$\theta_i^{\textnormal{action}} \in [0,1]$: action execution threshold of actor $a_i$
		\Statex \textbf{Inputs:}
		\Statex \hspace{0.3cm}$a_i$: actor
		\Statex \textbf{Output:}
		\Statex \hspace{0.3cm}Updated $\mathcal{H}_i^{\textnormal{action}}$
		\vspace{.2cm}
		\State $\mathcal{C}_i \leftarrow \textsc{Agent}_i(\cup_{j=1}^N\mathcal{H}_j^{post},\, \mathcal{H}_i^{\textnormal{action}},\, \rho_i)$ \Comment{Agent selects $M$ candidate posts for action}
		\For{each $c \in \mathcal{C}_i$}
		\State Draw $u \sim \mathcal{U}(0,1)$
		\If{$u > \theta_i^{\textnormal{action}}$}
		\State $\textsc{Vote}(c)$ \Comment{Upvote or downvote $c$ via $\mathcal{T}$}
		\State $\mathcal{H}_i^{\textnormal{action}} \leftarrow \mathcal{H}_i^{\textnormal{action}} \cup \{(a_i,\,c,\, \textsc{Vote}(c))\}$ \Comment{Update action history of $i$-th actor}
		\EndIf
		\EndFor
	\end{algorithmic}
\end{algorithm}

Algorithm~\ref{alg:post} presents the \textsc{Post} procedure, which governs the content generation behavior of actor $a_i$. At each posting event, the actor determines whether to submit a new comment or a reply by drawing $v \sim \mathcal{U}(0,1)$ and comparing against the reply probability $p_i^{\textnormal{reply}}$. A higher $p_i^{\textnormal{reply}}$ encodes a reactive communication style, wherein the actor predominantly responds to existing contributions rather than initiating 
independent arguments. In the reply case, the agent reasons over the post histories of all other actors $\bigcup_{j=1}^N \mathcal{H}_j^{\textnormal{post}}$ to select a target post $t^*$, either to agree or disagree with, consistent with its persona $\rho_i$. In both cases, the agent may optionally invoke the shared tool suite $\mathcal{T}$. For instance, performing web searches to incorporate references prior to generating content. The resulting post $c$ is published to the platform via $\mathcal{T}$ and recorded in $\mathcal{H}_i^{\textnormal{post}}$.

\begin{algorithm}[!ht]
	\caption{$\textsc{Post}(a_i)$}\label{alg:post}
	\begin{algorithmic}[1]
		\Statex \textbf{Parameters:}
		\Statex \hspace{0.3cm}$\rho_i$: Persona of the $i$-th actor $a_i$
		\Statex \hspace{0.3cm}$p_i^{\textnormal{reply}} \in [0,1]$: reply probability of actor $a_i$
		\Statex \textbf{Inputs:}
		\Statex \hspace{0.3cm}$a_i$: actor
		\Statex \textbf{Output:}
		\Statex \hspace{0.3cm}Updated $\mathcal{H}_i^{\textnormal{post}}$
		\vspace{.2cm}
		\State Draw $v \sim \mathcal{U}(0,1)$
		\If{$v \leq p_i^{\textnormal{reply}}$} \Comment{Reply with probability $p_i^{\textnormal{reply}}$}
		\State $t^* \leftarrow \textsc{Agent}_i(\cup_{j=1}^N\mathcal{H}_j^\textnormal{post},\, \rho_i)$ \Comment{Agent selects target to agree or disagree with}
		\State $c \leftarrow \textsc{Agent}_i(t^*,\, \mathcal{T})$ \Comment{Agent optionally invokes $\mathcal{T}$ to enrich reply}
		\Else \Comment{New comment with probability $1 - p_i^{\textnormal{reply}}$}
		\State $c \leftarrow \textsc{Agent}_i(\cup_{j=1}^N\mathcal{H}_j^\textnormal{post},\, \rho_i,\, \mathcal{T})$ \Comment{Agent optionally invokes $\mathcal{T}$ to enrich comment}
		\EndIf
		\State $\textsc{Publish}(c)$ \Comment{Post content to platform via $\mathcal{T}$}
		\State $\mathcal{H}_i^{\textnormal{post}} \leftarrow \mathcal{H}_i^{\textnormal{post}} \cup \{(a_i, c)\}$ \Comment{Update post history of $i$-th actor}
	\end{algorithmic}
\end{algorithm}

\section{Use Case Scenario: Integration with \textsc{Deliberate}}\label{Sec:5}

To validate \textsc{Chorus} in a realistic deployment setting, we integrated the framework with \textsc{Deliberate}, an AI-powered web platform for facilitating citizen engagement on policy issues through structured online discussions. \textsc{Deliberate} employs NLP-based pipelines to extract thematic trends, identify consensus and contention and generate deliberation reports from user contributions \citep{Livieris2026}, supporting evidence-based policymaking through synthesized stakeholder insights. This section describes the pilot deployment configuration, presents the generated discussion outputs and reports the results of an expert evaluation assessing content realism, discussion coherence and analytical utility.

The pilot scenario simulated a public deliberation titled ``\textit{Extreme Weather Events Due to Climate Change}''\footnote{The complete source code of \textsc{Chorus}, together with the discussion data generated during this pilot, are publicly available at \url{https://github.com/novelcore/chorus-simulation-framework/}}, instantiating $N=10$ actors across four persona archetypes over a simulation horizon of $T=20$ minutes. All agents are powered by Claude Sonnet 4.5, selected for its balance between generation quality and cost efficiency across sustained multi-actor simulations and  completed at an estimated API cost of less than 2 euros. The cast comprised four (4) Casual Users, one (1) Expert, three (3) Advocates and two (2) Skeptics, whose behavioral profiles are summarized as follows:

\begin{itemize}
	\item \textit{Casual User:} Polite, conversational participants with an informal tone and anecdotal, experience-driven content. They produce short sentence-level responses (10--20 words), operating with partially enabled discussion history (recent context only).
	\item \textit{Expert:} A polite, educational participant with a formal tone and methodical, evidence-backed content quality. Produces extended paragraph-level responses (50--100 words), leveraging full discussion history.
	\item \textit{Advocate:} Assertive, mission-driven participants with a semi-formal tone and policy-informed, advocacy-oriented content quality. They produce moderate sentence-to-paragraph-level responses (20--30 words), operating with partially enabled discussion history.
	\item \textit{Skeptic:} Pragmatic and analytical participants with a direct tone and scrutinizing, fact-focused content quality. They produce structured paragraph-level responses (30--50 words), leveraging full discussion history with active tracking of prior claims.
\end{itemize}

The archetype distribution approximates the stakeholder composition typical of real-world deliberation platforms, where citizen voices constitute the majority of contributions, domain experts provide lower-frequency but higher-density input, advocates introduce policy-level framing and skeptics promote deliberative diversity through critical counterpoints. The corresponding behavioral parameters governing each archetype's temporal and interactional dynamics are reported in Table~\ref{Table:Actor configuration}. All actors are equipped with a shared base tool suite $\mathcal{T}$ comprising post submission, discussion history retrieval and voting (upvote/downvote) actions. In addition, the Expert actor is provisioned with a web search tool, enabling evidence retrieval from external sources prior to content generation, consistent with its archetype's evidence-backed communication profile.

\begin{table}[!ht]
	\centering
	\small
	\renewcommand{\arraystretch}{1}
	\setlength{\tabcolsep}{10pt}
	\begin{tabular}{@{}l|l|cccc@{}}
		\toprule
		Actor & Archetype & $\lambda_i^{\textnormal{post}}$ & $\lambda_i^{\textnormal{action}}$ & $p_i^{\textnormal{reply}}$ & $\theta_i^{\textnormal{action}}$ \\
		\midrule
		Casual User 1  & \multirow{4}{*}{Casual User} & 1.0 & 1.4 & 0.45 & 0.35 \\
		Casual User 2  &  & 0.7 & 1.0 & 0.40 & 0.40 \\
		Casual User 3  &  & 1.2 & 1.5 & 0.50 & 0.30 \\
		Casual User 4  &  & 0.5 & 0.7 & 0.35 & 0.45 \\
		\midrule
		Expert  & Expert & 0.4 & 0.6 & 0.60 & 0.65 \\
		\midrule
		Advocate 1  & \multirow{3}{*}{Advocate} & 1.0 & 1.8 & 0.55 & 0.25 \\
		Advocate 2  &  & 1.2 & 2.0 & 0.60 & 0.20 \\
		Advocate 3  &  & 0.8 & 1.6 & 0.50 & 0.30 \\
		\midrule
		Skeptic 1  & \multirow{2}{*}{Skeptic} & 0.55 & 1.3 & 0.70 & 0.55 \\
		Skeptic 2  &  & 0.45 & 1.1 & 0.75 & 0.60 \\
		\bottomrule
	\end{tabular}
	\caption{Actor configuration and behavioral parameters. $\lambda_i^{\textnormal{post}}$ 
		and $\lambda_i^{\textnormal{action}}$ denote the expected number of posts and actions 
		per unit time, $p_i^{\textnormal{reply}}$ the probability of replying to an existing 
		post, and $\theta_i^{\textnormal{action}}$ the action selectivity threshold.}    \label{Table:Actor configuration}
\end{table}

\begin{figure}[!ht]
	\centering
	\includegraphics[width=0.95\linewidth]{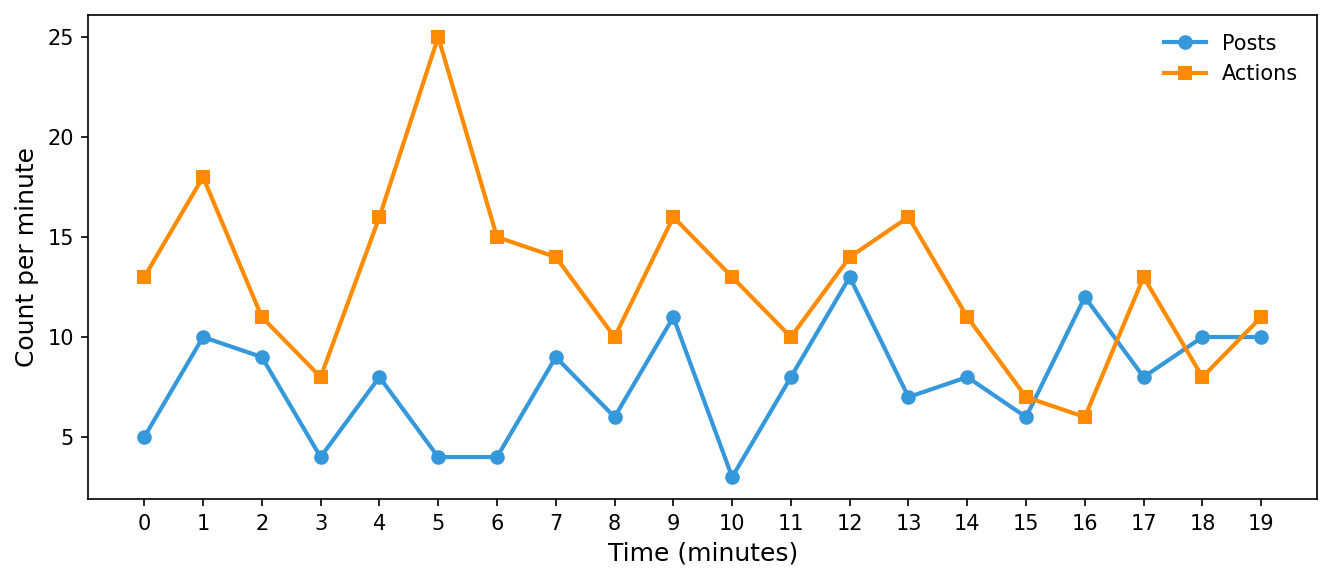}\\[6pt]
	\includegraphics[width=0.95\linewidth]{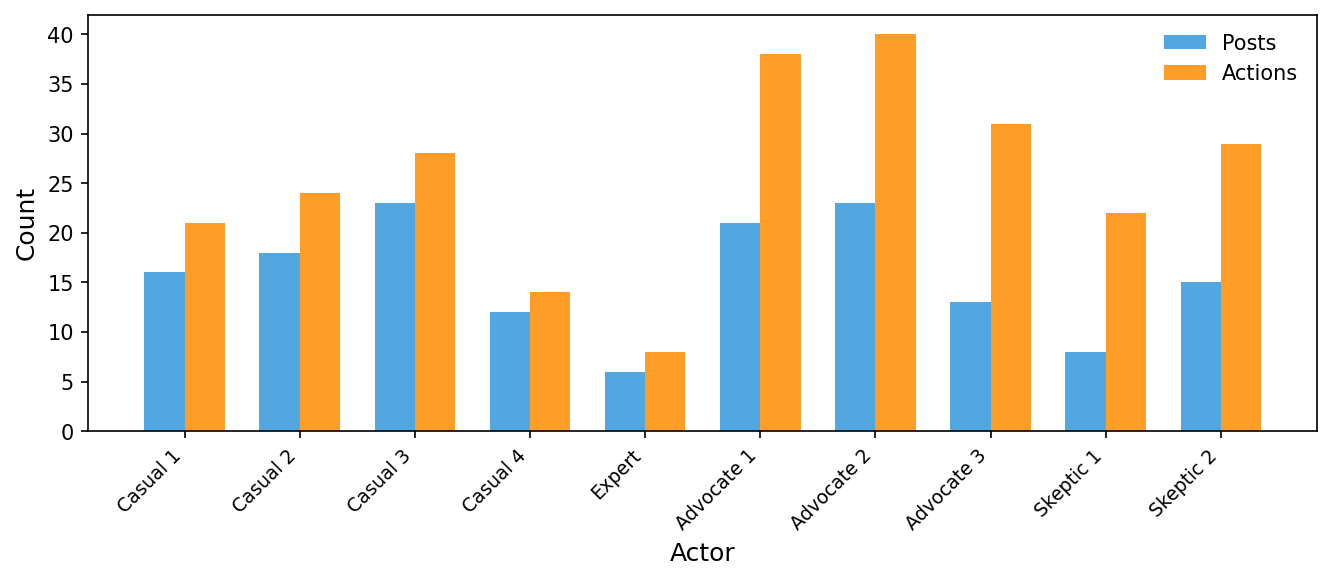}\\[6pt]
	\includegraphics[width=0.95\linewidth]{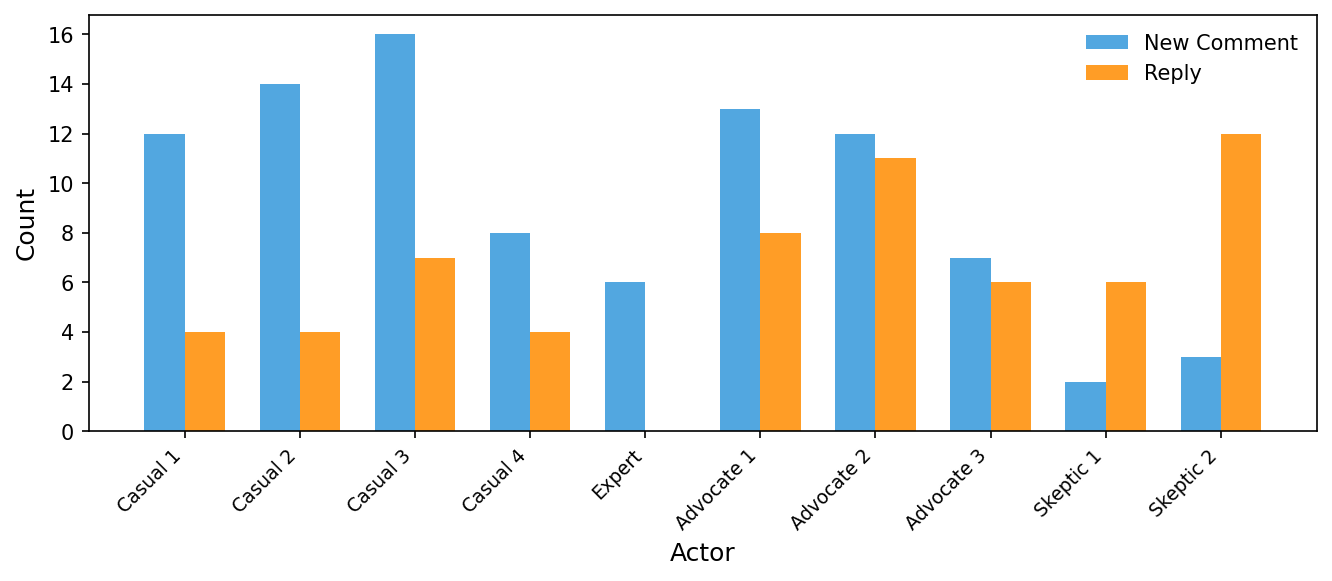}
	\caption{Simulated activity over $T=20$ minutes: (a) posts and actions per minute across all 
		actors, (b) aggregate post and action counts per actor and (c) new comments versus replies 
		per actor.}
	\label{Fig:Activity}
\end{figure}

Figure~\ref{Fig:Activity} illustrates the simulated activity produced by \textsc{Chorus} over the $T=20$ minute horizon, reporting per-minute participation dynamics, per-actor engagement volumes and the distribution of new comments versus replies. 

Figure~\ref{Fig:Activity}(a) confirms that the Poisson process-based temporal model produces sustained yet irregular participation, with posting rates fluctuating between 3 and 13 posts per minute and voting actions ranging from 6 to 25 per minute. Actions consistently outnumber posts throughout the simulation, reflecting the higher $\lambda^{\textnormal{action}}$ parameterization across all archetypes. Neither signal exhibits monotonic trends or periodic structure, approximating the bursty, non-stationary activity patterns characteristic of real online platforms.
Figure~\ref{Fig:Activity}(b) reveals the intended engagement asymmetry across archetypes. Actors: ``Advocate 1'' and ``Advocate 2'' are the most active contributors in both posts and actions, while the Expert actor produces the fewest posts consistent with its lower $\lambda^{\textnormal{post}}$ value. Notably, action counts are disproportionately high relative to post counts across all actors, with ``Skeptic 2'' exhibiting one of the largest action-to-post ratios despite moderate posting volume, reflecting a tendency to engage reactively with existing content rather than initiating new threads.
Finally, Figure~\ref{Fig:Activity}(c) validates the behavioral differentiation encoded in $p_i^{\textnormal{reply}}$. Casual Users produce predominantly new comments with few replies, consistent with their low reply probability of 0.30. Advocates exhibit a more balanced distribution, reflecting their intermediate 
$p^{\textnormal{reply}} = 0.50$. ``Skeptic 2'' is the most notable case, where replies outnumber new comments, consistent with the archetype's high $p^{\textnormal{reply}} = 0.75$ and its persona-driven tendency to critically engage with prior assertions rather than introduce independent arguments. The actor: ``Expert'' produces exclusively new comments, reflecting both its low reply probability and its role as an anchor of technical content within the discussion.

\textsc{Chorus} was preliminarily evaluated by a group of 30 experts involved in the \textsc{Deliberate} project, comprising QA practitioners, linguists and AI researchers overseeing NLP pipeline integration. The evaluation was conducted via an online questionnaire using a five-point Likert scale (1: strongly disagree, 2: disagree, 3: moderate, 4: agree, 5:strongly agree) across three dimensions: content realism, discussion coherence and analytical utility. The corresponding evaluation questions were formulated as follows: (i) \textit{``Does the generated synthetic content convincingly approximate real user participation in terms of linguistic quality and behavioral diversity?''}, (ii) \textit{``Do the simulated discussions exhibit coherent argumentation, diverse viewpoints and natural interaction patterns consistent with real deliberation?''} and (iii) \textit{``Does the discussion produce interpretable outputs when processed by the platform's NLP analytical pipelines?''}

Results were encouraging across all dimensions, as summarized in Table~\ref{tab:eval}. Content realism scored highest ($\mu = 4.6$), reflecting evaluator confidence that synthetic contributions were difficult to distinguish from genuine user input in terms of tone, vocabulary and argumentation style. Discussion coherence received a slightly lower score ($\mu = 4.1$), attributable to the inherent difficulty of replicating emergent multi-party discourse dynamics, including topic drift and implicit contextual references, through persona-conditioned generation alone. Analytical utility scored $\mu = 4.3$, confirming that \textsc{Chorus}-generated data constitutes a suitable input for downstream NLP pipelines, yielding thematic outputs consistent with genuine climate adaptation discourse.

\begin{table}[!ht]
	\centering
	\small
	\renewcommand{\arraystretch}{1.3}
	\begin{tabular}{@{}lcc@{}}
		\toprule
		\textbf{Dimension} & \textbf{Mean Score} & \textbf{Description} \\
		\midrule
		Content Realism      & 4.6 & Linguistic quality and behavioral diversity \\
		Discussion Coherence & 4.1 & Argumentation coherence and interaction patterns \\
		Analytical Utility   & 4.3 & Suitability for downstream NLP pipeline processing \\
		\bottomrule
	\end{tabular}
	\caption{Expert evaluation results across the three assessed dimensions (5-point Likert scale).}
	\label{tab:eval}
\end{table}

\section{Conclusions and Future Research}\label{Sec:6}

In this work, we presented an agentic-based framework named \textsc{Chorus},  which orchestrates LLM-powered actors with behaviorally consistent personas to generate realistic deliberation data on interactive web platforms. The framework addresses a critical bottleneck in online discourse analysis: the scarcity of large-scale deliberation data imposed by restrictive platform access policies, ethical constraints surrounding user privacy and inconsistent dataset quality. By grounding synthetic user behavior in structured personas, a principled Poisson process-based temporal model and structured tool usage, \textsc{Chorus} provides researchers and platform operators with a controlled yet richly diverse source of deliberation data suitable for discourse analysis, platform demonstration and NLP pipeline evaluation.

\textsc{Chorus} was deployed on the \textsc{Deliberate} platform within a climate adaptation pilot scenario, instantiating 10 actors across four persona archetypes over a 20-minute simulation horizon. The framework successfully produced a heterogeneous discussion exhibiting the engagement asymmetry, topical diversity and interaction patterns characteristic of authentic participatory discourse. Preliminary expert evaluation across 30 participants yielded encouraging results relative to content realism, analytical utility and discussion coherence, confirming that \textsc{Chorus}-generated data constitutes a credible substitute for real user participation in platform demonstration and analytical pipeline validation.

A key limitation of the current evaluation concerns the absence of ablation studies isolating the contribution of individual framework components. Comparisons against simpler baselines, such as a single LLM generating the complete discussion monolithically, or configurations omitting persona-driven behavioral grounding or Poisson process-based temporal modeling, would provide direct quantitative evidence for the necessity of each design decision. Addressing this gap constitutes an important direction for future work. Moreover, the current evaluation does not quantify tool invocation frequency or its effect on content quality and relies exclusively on Claude Sonnet 4.5, leaving the empirical contribution of tool usage and the sensitivity to model substitution unexamined. A systematic analysis of tool utilization patterns and cross-model evaluation are planned as part of the framework's forthcoming extension.

In our future work, we intend to extend the current capabilities of \textsc{Chorus} by  involving adversarial simulation, where specialized archetypes introduce disruptive behaviors such as polarizing rhetoric and deliberate misinformation, enabling the study of platform resilience under controlled conditions. A second direction concerns adaptive actor archetypes, wherein  behavioral strategies are iteratively refined in response to the evolving discussion state, moving beyond static persona parameterization toward dynamic agent adaptation. Finally, we intend to conduct larger-scale validation studies involving comparative evaluation against real deliberation data and deployment across diverse policy domains, with the long-term goal of establishing \textsc{Chorus} as a robust benchmarking tool for participatory platform research.

\noindent \textbf{Acknowledgements} This work received funding from the Horizon Europe research and innovation programme under Grant Agreement No. 101137711, project NEUROCLIMA (Developing and assessing novel educational and user-centred actions towards scaling up behavioural change and climate resilience through an AI-enhanced solution).

\clearpage

\appendix

\section{Use Case Outputs: Thematic Trend Analysis}

The thematic clusters presented below are produced by \textsc{Deliberate}'s automated NLP analysis pipeline and are included to demonstrate the analytical utility of \textsc{Chorus}-generated data: the ability of the framework to produce discussions of sufficient thematic richness and structural diversity to yield interpretable, policy-relevant outputs without reliance on real user participation.\\

\noindent{}\textbf{Trend 1}
\begin{itemize}
	\item[--] \textit{Title}: Immediate Cooling Solutions for Vulnerable Populations During Extreme Heat Events.
	\item[--] \textit{Summary}: Vulnerable individuals, particularly the elderly, face life-threatening conditions during extreme heat yet cannot afford cooling systems. Existing air-conditioned public buildings should be immediately opened as cooling centers, with accessible information on their locations to prevent health crises.
	\item[--] \textit{Keywords}: ``summer'', ``cooling'', ``hospital'', ``apartment'', ``buildings'', ``people'', ``years'', ``day'', ``numbers'', ``construction''.
\end{itemize}

\noindent{}\textbf{Trend 2}
\begin{itemize}
	\item[--] \textit{Title}: Equitable Climate Adaptation: Addressing Infrastructure Needs and Funding Gaps.
	\item[--] \textit{Summary}: Climate adaptation funding disproportionately favors those with existing resources, while complex EU funding processes further limit uptake by under-resourced municipalities. The absence of enforceable standards mandating equity assessments and accessible thermal refuges represents a critical policy gap.
	\item[--] \textit{Keywords}: ``adaptation'', ``cooling'', ``climate'', ``costs'', ``infrastructure'', ``billion'', ``thermal'', ``heat events'', ``requirements'', ``policy''.
\end{itemize}

\noindent{}\textbf{Trend 3}
\begin{itemize}
	\item[--] \textit{Title}: Supporting Agricultural Resilience Amidst Climate Disasters: The Overlooked Crisis.
	\item[--] \textit{Summary}: Agricultural workers and rural communities in Greece remain overlooked amid urban-focused discussions, with drought and fires decimating farms and threatening food security. Without targeted support, insurance mechanisms and planning priorities beyond urban cooling, Greece risks irreversible decline in agricultural self-sufficiency.
	\item[--] \textit{Keywords}: ``coastal'', ``infrastructure'', ``building codes'', ``cooling'', ``cooling systems'', ``water'', ``food'', ``heat''.
\end{itemize}

\noindent{}\textbf{Trend 4}
\begin{itemize}
	\item[--] \textit{Title}: Healthcare Infrastructure and Preparedness: Addressing the Challenges of Extreme Heat Events.
	\item[--] \textit{Summary}: Extreme heat events are overwhelming emergency services and exposing failures in healthcare preparedness, with under-cooled neighborhoods recording significantly higher admission rates. A comprehensive approach integrating phased retrofitting, updated regulatory standards and climate-resilient urban design is needed to reduce heat-related mortality.
	\item[--] \textit{Keywords}: ``infrastructure'', ``heat'', ``cooling'', ``thermal'', ``buildings'', ``emergency'', ``costs'', ``healthcare'', ``social housing'', ``climate'', ``performance'', ``housing'', ``capacity'', ``temperatures''.
\end{itemize}

%%%%%%%%%%%%%%%%%%%%%%%%%%%%%%%%%%
\bibliographystyle{unsrtnat}
\bibliography{bibliography}
%%%%%%%%%%%%%%%%%%%%%%%%%%%%%%%%%%

\end{document}